\documentclass[letterpaper, 10 pt, conference]{ieeeconf}  
 
\usepackage{cite}
\usepackage{multicol}
\usepackage[bookmarks=true]{hyperref}
\usepackage{algpseudocode}
\usepackage{algorithm}
\usepackage{amsmath}
\usepackage[dvipsnames]{xcolor}
\usepackage{multirow}
\usepackage{ctable}
\usepackage{caption}
\usepackage{subcaption}
\usepackage{ragged2e}
\usepackage{amsfonts}
\usepackage{comment}

\usepackage{fancyhdr}
\hypersetup{
    colorlinks=true,
    linkcolor=blue,
    filecolor=magenta,      
    urlcolor=cyan,
    pdftitle={Overleaf Example},
    pdfpagemode=FullScreen,
    }

\IEEEoverridecommandlockouts                              

\title{\LARGE \bf
Self-supervised Modular Neural Learning for Robot Navigation in Unknown, Large Indoor Environments}

\title{\LARGE \bf
Online Hierarchical Policy Learning using Physics Priors for Robot Navigation in Unknown Environments}

\author{Wei Han Chen*, Yuchen Liu*, Alexiy Buynitsky, and Ahmed H. Qureshi
\thanks{Wei Han Chen, Yuchen Liu, Alexiy Buynitsky, and Ahmed H. Qureshi are with the Department of Computer Science, Purdue University, West Lafayette, IN, USA, 47907. Email {\tt\small$\{$chen3189, liu3853, abuynits, ahqureshi$\}@$purdue.edu}}%
\thanks{* denotes equal contribution}
}

\begin{document}

\let\oldtwocolumn\twocolumn
\renewcommand\twocolumn[1][]{%
    \oldtwocolumn[{#1}{
    \begin{center}
    \vspace{-5mm}
    \includegraphics[width=\textwidth]{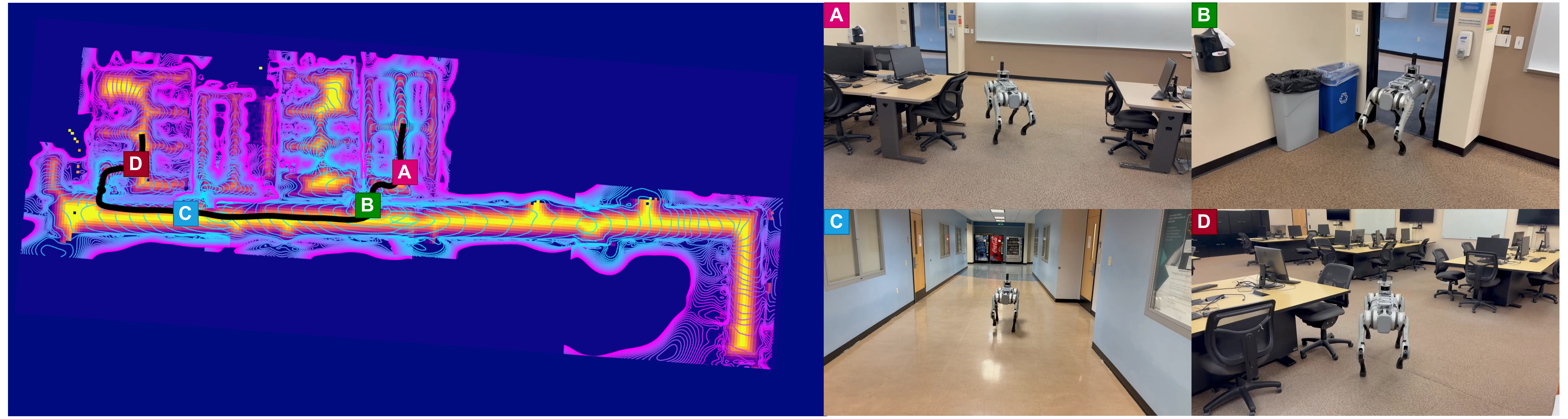}
    \captionof{figure}{\justifying We demonstrate our approach in a real-world indoor scenario. The robot begins in the rightmost room, adjacent to an obstacle, and navigates through multiple doorways and narrow passages to reach its goal in the leftmost room. The environment is segmented into 6 subnetworks, with local travel times visualized as contour lines. In this particular trial, the entire path planning process was completed in 0.08 seconds.}
           \label{fig:lwsn_basement}
        \end{center}
    }]
}

\maketitle
\thispagestyle{empty}
\pagestyle{empty}

\begin{abstract}

Robot navigation in large, complex, and unknown indoor environments is a challenging problem. The existing approaches, such as traditional sampling-based methods, struggle with resolution control and scalability, while imitation learning-based methods require a large amount of demonstration data. Active Neural Time Fields (ANTFields) have recently emerged as a promising solution by using local observations to learn cost-to-go functions without relying on demonstrations. Despite their potential, these methods are hampered by challenges such as spectral bias and catastrophic forgetting, which diminish their effectiveness in complex scenarios. To address these issues, our approach decomposes the planning problem into a hierarchical structure. At the high level, a sparse graph captures the environment’s global connectivity, while at the low level, a planner based on neural fields navigates local obstacles by solving the Eikonal PDE. This physics-informed strategy overcomes common pitfalls like spectral bias and neural field fitting difficulties, resulting in a smooth and precise representation of the cost landscape. We validate our framework in large-scale environments, demonstrating its enhanced adaptability and precision compared to previous methods, and highlighting its potential for online exploration, mapping, and real-world navigation. \href{https://sites.google.com/view/mntfields/home}{https://sites.google.com/view/mntfields/home}

\end{abstract}

\section{Introduction}

Navigating large, unknown environments presents significant challenges in robotics, where both mapping and planning are crucial yet difficult tasks. Traditional mapping approaches often generate occupancy or signed distance field (SDF) maps that require additional processing, such as grid search or optimization, to extract navigable paths \cite{vasilopoulos2024hio, Reijgwart_2020, valero2013path, 4082128, zucker2013chomp}. This extra step not only increases computational overhead but also complicates the transition from a raw map to an actionable navigation plan.

Alternatively, some methods directly build probabilistic roadmaps (PRMs) from sensor data \cite{kavraki1996probabilistic, faust2018prm}. While these techniques can yield efficient paths in simpler settings, they tend to become unwieldy in expansive, complex environments. The sheer number of nodes required to accurately represent intricate spaces often leads to memory-intensive computations and difficulties in maintaining and controlling the roadmap structure over large scales.

More recently, methods like Neural Time Fields (NTFields) \cite{ni2023ntfields} have been introduced to infer cost-to-go functions directly by solving the Eikonal equation. These approaches aim to bypass the intermediate mapping step by providing an implicit representation of the navigation cost. However, NTFields encounter significant hurdles when scaled up to large scenes. Their reliance on neural network architectures introduces issues such as spectral bias, catastrophic forgetting, and poor convergence. Moreover, the inherent scaling challenge of solving a partial differential equation (PDE) further complicates their application in diverse, cluttered environments.

Inspired by the hierarchical planning strategies we use in everyday navigation—such as how mapping applications outline broad routes while vehicles make local maneuvers—we propose a modular, hierarchical approach called Modular-NTFields (mNTFields) to address navigation challenges. 

At a high level, our method constructs an online sparse navigation graph from local observations, capturing the connectivity between different subparts of a large environment. This high-level strategy is informed by our low-level method and yields a compact and efficient representation, allowing for rapid global planning without the burden of excessive detail. On a local level, we integrate NTFields to develop detailed cost-to-go maps for each subpart based on local observations. By solving the Eikonal equation locally, our approach effectively manages obstacles and complex geometries while alleviating the scalability issues that often affect traditional NTFields. Additionally, we enhance local planning with Temporal Difference Metric Learning (TDM) \cite{ni2025physicsinformed} to further improve convergence to an accurate Eikonal PDE solution.

We validate mNTFields across several challenging scenarios, demonstrating that it outperforms existing methods—particularly in large, complex, unknown indoor environments where standard approaches often struggle. Our experiments show that mNTFields achieves significantly faster navigation with higher success rates. We also showcase its practical deployment on a quadruped robot navigating through multiple rooms and narrow passages. This work highlights the potential of hierarchical planning frameworks for advancing robust and scalable robot navigation in unknown environments.

\section{Related Work}

Efficient navigation in partially or fully unknown environments remains a fundamental challenge in robotics, often requiring reliance on local sensor observations \cite{982903}. Traditional motion planning methods include sampling-based approaches \cite{lavalle2001rapidly, xu2021autonomous, kavraki1996probabilistic}, which iteratively build a roadmap of feasible paths, and grid-based methods \cite{valero2013path, 4082128}, which discretize the environment into occupancy maps for search algorithms. While these techniques perform well in lower-dimensional or structured environments, their reliance on explicitly maintaining large graphs or grids makes them computationally expensive in complex, high-dimensional spaces. In contrast, our hierarchical representation enables efficient pathfinding for arbitrary start-goal pairs without requiring exhaustive search over large maps.

Recent data-driven approaches such as reinforcement learning (RL) \cite{Gao2020Deep, Marchesini2020DiscreteDR, Liu2024DeepRL} and imitation learning (IL) \cite{ichter2018learning, qureshi2019motion, qureshi2018deeply, qureshi2020motion} attempt to directly learn control policies from high-dimensional sensor inputs (e.g., images), often leveraging photorealistic simulations. However, these methods typically require extensive interaction data or expert demonstrations, which can be costly to obtain for real-world applications. Our method sidesteps this issue by leveraging physics-based principles, reducing the dependency on large expert datasets and improving adaptability across different environments.

Another class of solutions focuses on ego-centric local planning \cite{yang2023iplanner, Roth2023ViPlannerVS}, where navigation is based solely on onboard sensing, without global information or expert supervision. These approaches are well-suited for dynamic environments as they continuously update their plans based on new observations. However, their reliance on local information often leads to suboptimal routes or getting trapped in local minima. By maintaining a hierarchical global representation, our approach preserves adaptability while incorporating long-range planning, reducing the risk of poor local decisions.

More recently, foundation models \cite{Shah2023ViNT:, Zhang2024Vision-and-Language}, trained on large-scale datasets from expert demonstrations or powerful planners, have shown promising generalization across diverse environments. However, their heavy computational demands and reliance on vast amounts of high-quality training data limit their scalability. These models may also struggle in highly complex or novel environments where their pre-trained knowledge does not generalize well. In contrast, our approach offers a lightweight, scalable solution that can be efficiently deployed in large, complex workspaces without requiring extensive pre-training.

A growing body of work explores physics-informed neural networks \cite{ni2023ntfields, smith2020eikonet, ni2023progressive, ni2024physics, ni2025physicsinformed}, which embed physical constraints into neural networks to guide learning without expert data. Methods like ANTFields \cite{antfield} have demonstrated efficiency in moderately complex environments, but their scalability is hindered by challenges such as the complex loss landscape imposed by PDE constraints and the spectral bias of neural networks, which makes it difficult to capture high-frequency variations in large environments \cite{moseley2023finite}. 

To overcome these scalability limitations, our approach builds upon the online training pipeline of ANTFields and the efficiency of TDM \cite{ni2025physicsinformed} while introducing a hierarchical structure that systematically decomposes the environment into smaller, more manageable submodules. This design extends the applicability of physics-informed learning to significantly larger and more intricate environments, offering a balance between computational efficiency and global planning performance.

\section{Background}

This section provides an overview of mapping and motion planning in robotics, with a particular focus on physics-informed neural motion planners. 

Let the robot’s $d$-dimensional workspace be denoted by $\mathcal{X} \in \mathbb{R}^d$ and its $m$-dimensional configuration space (C-space) by $\mathcal{Q} \in \mathbb{R}^m$. The obstacle regions are defined as $\mathcal{X}_{\mathrm{obs}}$ in the workspace and $\mathcal{Q}_{\mathrm{obs}}$ in the C-space, while the corresponding free spaces are $\mathcal{X}_{\mathrm{free}}$ and $\mathcal{Q}_{\mathrm{free}}$, respectively. our key objective is to navigate unknown environments by building their navigation-friendly, reusable, and compact representations $\mathcal{F}$ via exploration related to $\mathcal{X}_{\mathrm{obs}}$.

One classical approach to motion planning is to leverage the Eikonal Equation. In this formulation, the arrival time function \(T(q_s, q_g)\) between a start configuration \(q_s\) and goal configuration \(q_g\) satisfies

\begin{equation}
\frac{1}{S(q_g)} = \|\nabla_{q_g} T(q_s,q_g)\|
\label{eikonal}
\end{equation}

where the speed function \(S(q)\) is designed to decrease near obstacles and increase in free space. This relationship provides a basis for constructing a cost map that naturally encodes collision avoidance.

Building on this foundation, recent advances have integrated the Eikonal equation within the framework of physics-informed neural networks (PINNs) to enable neural motion planning. NTFields \cite{ni2023ntfields} is the first method within this paradigm. Subsequent variants have introduced enhancements: P-NTFields \cite{ni2023progressive} augments the formulation with a viscosity term, resulting in a semi-definite equation that improves stability, while Temporal Difference Metric learning (TDM) incorporates a temporal difference loss via a Taylor expansion of the arrival time function \cite{ni2025physicsinformed}. This additional loss term, along with auxiliary losses regarding obstacle-aware normal alignment and causality weighting, enables TDM to handle more complex environments effectively.

While these approaches were originally developed for known environments, Active-NTFields (ANTFields) \cite{antfield} extends these ideas to unknown settings by constructing a cost map from local depth observations. In ANTFields, the arrival travel time fields serve as a cost-to-go map feature $\mathcal{F}$, and their gradients inherently yield collision-free paths. Despite their rapid planning times and the advantage of not requiring expert data, these methods tend to suffer from spectral bias and difficulty in capturing high-frequency details when environments become increasingly complex.

To address these limitations, we propose a hierarchical structure that better maps and navigates unknown environments. In our framework, physics-informed neural networks are deployed over small spatial regions at a low level, while a high-level connectivity graph captures the broader environmental structure. Our low-level planning is inspired by ANTFields \cite{antfield}; however, we also augment its training with TDM \cite{ni2025physicsinformed} objective functions, which demand online approximation of surface normals for obstacle alignment loss. Finally, our two-tiered approach leverages the strengths of local PINN-based planning while overcoming the high-frequency and spectral bias issues encountered in complex scenarios.


\begin{figure*}[t]
\centering
\vspace{1.5mm}
    \includegraphics[width=1.0\textwidth]{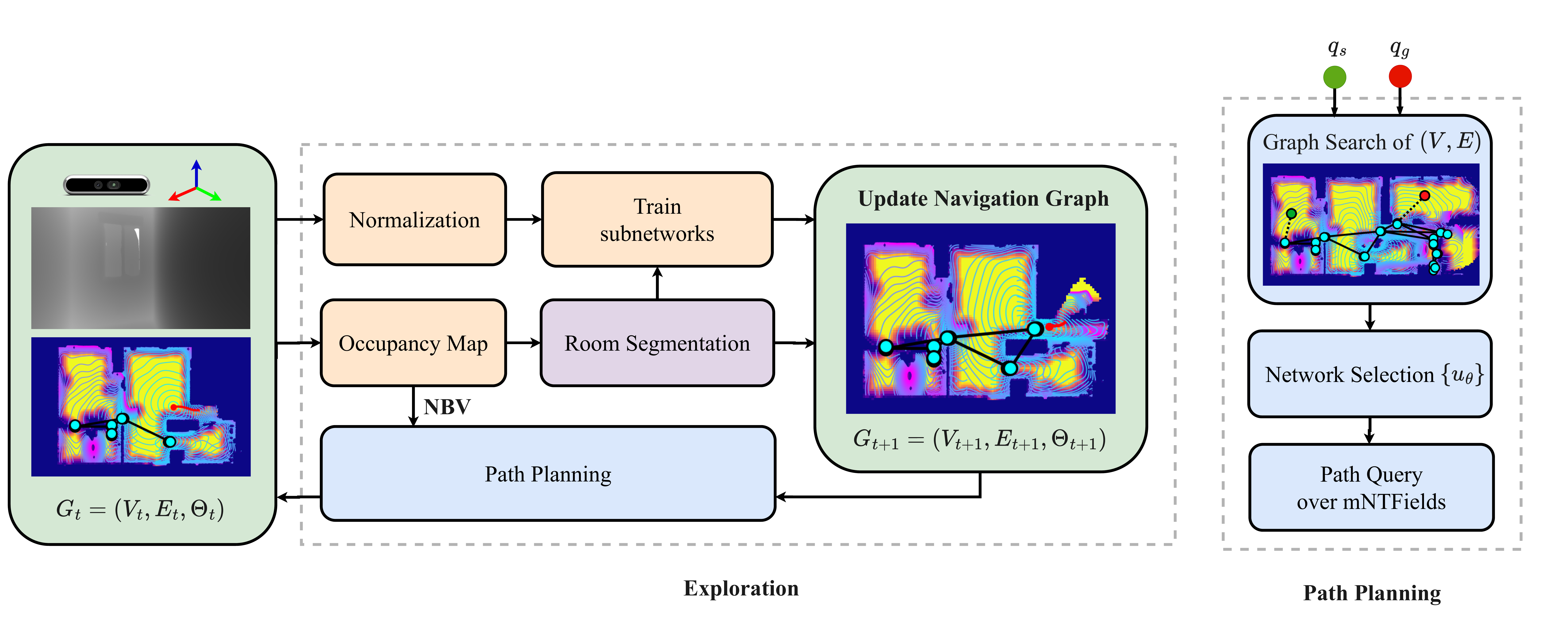}
    \caption{We propose mNTFields, a modular neural learning framework for scalable motion planning. Our pipeline constructs a navigation graph online during the exploration phase, which can then be leveraged for long-horizon path planning. The exploration phase begins with processing a local depth observation to build a global occupancy map. Then, room segmentation is performed to create new nodes in the navigation graph, where each node corresponds to a modular subnetwork. These subnetworks are trained using the normalized observation data. Finally, a path is planned towards the next best viewpoint to facilitate further exploration. During the path planning phase, a graph search is performed on the navigation graph. The corresponding subnetworks are queried to generate path segments, which are then concatenated to construct a full long-horizon path.}
    \label{fig:pipeline}
\vspace{-4.5mm}
\end{figure*}

\section{Proposed Methods}

This section presents Modular-NTFields (mNTFields), a hierarchical planning framework (Fig. \ref{fig:pipeline}) that segments indoor environments into parts using local observations. At a high level, it constructs a sparse graph capturing global connectivity. At a low level, multiple TDM \cite{ni2025physicsinformed} networks generate arrival time field maps for obstacle-aware navigation. Finally, we introduce a navigation approach that efficiently queries the high-level graph and local neural time fields for fast, goal-directed motion planning.

\subsection{High-Level Planning}

High-level path planning involves constructing a sparse graph that guides the assignment of low-level path planning to manageable regions. This is achieved through two key components: online room segmentation and navigation graph construction. Online room segmentation partitions the environment into distinct areas based on local observations, effectively isolating regions. Next, a navigation graph is built where well-explored rooms contribute entry points as nodes, and connectivity is established through edges weighted by predicted travel times based on our local planner. Together, these components enable efficient and scalable path planning in complex environments.

\subsubsection{Online Room Segmentation}

The first step is to partition the global occupancy map \(\mathcal{M}\) into manageable regions. This map is generated from local observations, including depth perception and odometry. Each cell in occupancy map \(\mathcal{M}\) is classified as free space, occupied, or unexplored. Room segmentation is performed using morphological segmentation \cite{morph}. A Morphological erosion operation is used to expand walls and seal narrow openings, such as doorways. Candidate rooms, denoted by \(\mathcal{R}\), are then obtained as the connected components of free space. We further subdivide \(\mathcal{R}\) into confirmed rooms, \(\mathcal{R}_c\subseteq\mathcal{R}\), and unconfirmed rooms, \(\mathcal{R}_u\subseteq\mathcal{R}\); a candidate room is assigned to \(\mathcal{R}_c\) if none of its boundary points adjoin unexplored areas. All other rooms are assigned to $\mathcal{R}_u$ Once a room is confirmed, it is treated as a static entity in subsequent segmentation iterations and serves as a stable training domain for the neural subnetworks. This procedure ensures that confirmed rooms remain unchanged and preserves connectivity at entry points between adjacent rooms.

\subsubsection{Navigation Graph Construction}

A global navigation graph \(G = (V, E, \Theta)\) is constructed as new rooms and doorways are discovered. Each confirmed room \(R_c\in\mathcal{R}\) contributes its entry points as nodes \(V\) and neural networks \(\Theta\) in the graph. For entry points within the same room, edges \(E\) are assigned weights corresponding to the travel times predicted by the neural time fields. In the case of adjacent rooms sharing an entry point, the corresponding nodes are connected with an edge of zero cost. This graph represents the spatial connectivity of the environment in a high level, which will be combined with low-level path planning in Section IV.B.

Our neural motion planning leverages modular networks to localize learning and mitigate spectral bias. Each confirmed room \( R_c \) is assigned to a subnetwork \( u_\theta \), which can be an existing subnetwork to prevent the generation of redundant networks. A room $R_c$ is assigned to an existing network $u_\theta$ if the following condition holds: (1) the room has sufficient overlap with the bounding box of $u_\theta$, (2) the room is connected to at least one other room already assigned to $u_\theta$, and (3) the resulting new bounding box of $u_\theta$ does not exceed a size threshold. if one of the conditions fails to hold, then a new network is instantiated and trained. This modularization ensures that training is localized, thereby reducing interference, expediting convergence, and enabling scalability in large environments.

\subsection{Low-Level Planning}

Our approach for low-level neural motion planning is inspired by ANTFields\cite{antfield}: we require configuration samples \(q_s, q_g\) paired with ground truth speed values \(S^*\) and obstacle surface normals \(\mathrm{N}^*\) to employ TDM objective functions. In the remainder of this section, we provide details of each component of our low-level pipeline.

\subsubsection{Sample Generation}
Adaptive data sampling and training for mNTFields are achieved by refining the stratified sampling method used in ANTFields to improve sampling data quality. In ANTFields, the samples are generated along the depth rays, and start and goal pairs are formed randomly from those samples. This results in an overrepresentation of free-space points and an unfavorable distance distribution between the start and goal pairs. 

In our adaptive sampling, we begin by generating start and goal points, \( q_s \) and \( q_g \), along with their closest obstacle points \( \bar{q}_s \) and \( \bar{q}_g \), similar to ANTFields approach. Next, to focus on more informative regions, only those start points that are close to obstacles (\( \|q - \bar{q}\| < d_{\max} \)) are retained. Distances are then sampled from a normal distribution \( d \sim \mathcal{N}(0,\sigma^2) \) with standard deviation \(\sigma\) and angles \(\alpha\) from a uniform distribution over \([0, 2\pi)\) to produce initial candidate goal points \( q_g = q_s + d \cdot (\cos\alpha, \sin\alpha) \). The final goal points are determined by selecting the nearest points from $q_g$ using a KDTree \cite{grandits_geasi_2021}. 

\begin{figure*}[t]
\centering
\vspace{2.5mm}
    \includegraphics[width=1.0\textwidth]{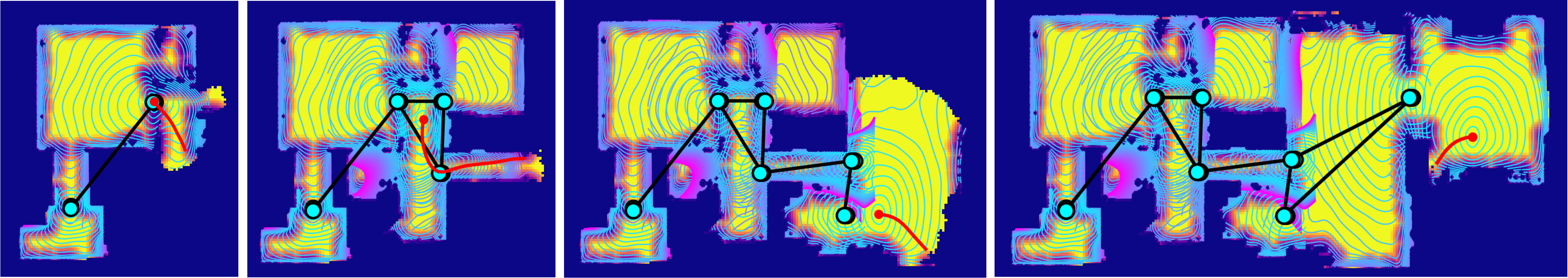}
    \caption{In the online exploration phase, new rooms are discovered with room segmentation. The entry points are identified and added to the graph, shown as the nodes in the figure. Entry points in the same room are also interconnected. To showcase the modular nature of our method, the travel time fields (cyan contour lines) are generated by separate networks. During exploration, we can use this graph along with the corresponding subnetworks to plan long horizon task with more robustness. The red dot shows the robot's current location, and the red lines shows the predicted trajectory to reach the next waypoint.}
    \label{fig:exploration}
\vspace{-4.5mm}
\end{figure*}

\subsubsection{Neural Architecture and Training Objectives}
Each subnetwork \(u_\theta \in \Theta\) is responsible for a specific room or set of spatially adjacent regions. In each subnetwork, all configurations are normalized within the subnetwork’s bounding box via 
\begin{equation}
q^* = \frac{q - b^{\min}}{b^{\max} - b^{\min}} - 0.5,
\label{normalization}
\end{equation}
where $b_{min}$ and $b_{max}$ are the minimum and maximum coordinates of the network's bounding box. Therefore, each subnetwork \(u_\theta\) receives normalized configuration start and goal samples \(q^*_s, q^*_g\) as input and outputs the predicted travel time \(T(q^*_s, q^*_g)\).

To train the subnetwork, the ground truth speeds, denoted by \((S^*(q^*_s),S^*(q^*_g))\), are computed as the Euclidean distance between configurations $(q^*_s, q^*_g)$ and their nearest obstacle points $(\bar{q}^*_s, \bar{q}^*_g)$ in normalized space. To employ the TDM objective function, we also need to approximate surface normals from local observation. We approximate these normals as:  
\(
\mathrm{N}^*(q^*) = \frac{\bar{q}^* - q^*}{\|\bar{q}^* - q^*\|}
\). Finally, inspired by TDM, we use the generated samples, their ground truth speed, and surface normals to define the training objective as follows:

The Eikonal loss is derived from Eq.~\ref{eikonal}:
\begin{equation}
L_E \;=\; \Bigl(\sqrt{\frac{S^*(q_s^*)}{S(q_s^*)}} - 1\Bigr)^2 + \Bigl(\sqrt{\frac{S^*(q_g^*)}{S(q_g^*)}} - 1\Bigr)^2,
\label{eikonal_loss}
\end{equation}

where $S(\cdot)$ is a predicted speed based on the Eikonal PDE (Eq.~\ref{eikonal}) governed by the gradient of neural predicted time field $T$. The TDM also enhances this loss with additional temporal difference and auxiliary losses. By performing a Taylor expansion of the value function \( T \) along the optimal policy \( \pi^* \) over a small time step \(\delta t\), the TD loss is defined as
\begin{equation}
\begin{aligned}
L_{\text{TD}} =\; &\Bigl[
T(q_s^*, q_g^*) 
- \frac{\delta t}{S^*(q_g^*)} 
- T\bigl(q_s^*,\; q_g^* + \pi_g^*\,\delta t\bigr)
\Bigr]^2 \\
+\;&\Bigl[
T(q_s^*, q_g^*) 
- \frac{\delta t}{S^*(q_s^*)} 
- T\bigl(q_s^* + \pi_s^*\,\delta t,\; q_g^*\bigr)
\Bigr]^2,
\end{aligned}
\label{td_loss}
\end{equation}
where the optimal policy is given by $\pi_g^* \;=\; -\frac{\nabla_{q_g^*} T(q_s^*, q_g^*)}{\|\nabla_{q_g^*} T(q_s^*, q_g^*)\|}$ and a similar expression for \( \pi_s^* \). This formulation promotes value propagation that aligns with the optimal policy. To further enhance obstacle avoidance, TDM employs an obstacle-aware normal alignment loss:
\begin{equation}
\begin{aligned}
L_N =\; &\bigl(1 - S^*(q_s^*)\bigr)
\Bigl\|\,
S^*(q_s^*)\,\nabla_{q_s^*} T(q_s^*, q_g^*)
+ \mathrm{N}^*(q_s^*)
\Bigr\|^2 \\
+\;&\bigl(1 - S^*(q_g^*)\bigr)
\Bigl\|\,
S^*(q_g^*)\,\nabla_{q_g^*} T(q_s^*, q_g^*)
+\mathrm{N}^*(q_g^*)
\Bigr\|^2.
\end{aligned}
\label{normal_loss}
\end{equation}
Here, the normalized gradients represent the ground-truth normal directions in C-Space, and the factor \( (1 - S^*(\cdot)) \) activates the loss only near obstacles. The complete loss to train our function is defined as follows:
\begin{equation}
L \;=\; \Bigl(\lambda_E\,L_E + \lambda_{\text{TD}}\,L_{\text{TD}} + \lambda_N\,L_N\Bigr)\,L_C,
\label{total_loss}
\end{equation}
where $L_C$ is a causality weight that prioritizes regions with lower arrival times. The \(\lambda_E, \lambda_{\text{TD}}, \lambda_N\) are user-defined hyperparameters. For more details on the objective function, please refer to \cite{ni2025physicsinformed}. 



Our neural architecture are also based on TDM's neural structure \cite{ni2025physicsinformed}. It comprises a configuration encoder $f$ and a metric function $D$ that computes the geodesic distance between latent encodings of start and goal configurations. The encoder \(f(q^*)\) = \(g(\gamma(q^*))\) takes the normalized configuration and extracts their high-frequency Fourier features using function $\gamma$. These features are further processed by the PirateNets structure \(g(\cdot)\), which integrates a modified Multi-layer Perceptron (MLP) with a residual gating mechanism for improved performance and stability. Finally, the geodesic distance is computed as \(T(q^*_s, q^*_g) = D(f(q^*_s), f(q^*_g))\), where $D$ is a metric function based on $L_1$ and $L_\infty$ distance. This design enables our model to effectively approximate Eikonal solutions in unseen environments while preserving the desired metric properties of Eikonal PDE solution.

\subsection{Online Training of mNTFields via Exploration}

Our integrated framework, termed mNTFields, employs continuous online training during exploration. At each exploration step, the robot collects new sensor data, updates the occupancy map, and re-segments the environment to discover new rooms. The navigation graph is updated accordingly, and each subnetwork \(u_\theta\) containing unconfirmed rooms is incrementally trained with the latest data. Our exploration policy follows that of ANTFields with selection for the next best viewpoint based on the frontier approach. Frontier points are identified as boundaries between explored and unexplored regions, which are then clustered with DBSCAN \cite{dbscan}. The center of the cluster with the least estimated travel time is chosen to be the next viewpoint. An important augmentation in our approach is a backtracking mechanism: previous methods ANTFields plan paths to the next best view based solely on the shortest travel time, often targeting areas outside of the well-trained regions. In contrast, if the planned path to a newly observed best view is deemed unreachable due to exceeding fixed steps of navigation caused by insufficient training, a backtracking point will be selected as a new next best view. The backtracking point is the previous robot location from which that previous unreachable next best view was observed. Traveling to this point provides easier access to the waypoint by leveraging reliable, previously acquired training data, thereby improving exploration performance and reducing the risk of collisions. After traveling to this backtracking point, the robot will continue to travel to the previously unreachable next best view in the following exploration step. The exploration phase repeats the above steps until there are no more frontier points or all frontier points are deemed as noise by DBSCAN. An illustration of this exploration phase is shown in Fig. \ref{fig:pipeline} and Fig. \ref{fig:exploration}.

\subsection{Robot Navigation using mNTFields}
Navigation is achieved by utilizing both the high-level navigation graph and a set of low-level trained subnetworks. The navigation graph represents the connections between confirmed rooms through their entry points, while the neural time fields provide estimations of local travel times. When a navigation request is made, specifying a start and goal configuration, we identify the rooms that these points reside in, connecting them to all entry points of their respective rooms. A graph search using Dijkstra's algorithm is then performed to retrieve the shortest sequence of nodes between the specified start and goal. The edges of our high-level graph are encoded with travel times, allowing the graph search to determine the shortest sequence. This results in a high-level path that connects the start to each entry point leading to the goal. For each consecutive node in the high-level path, the corresponding subnetwork computes a local path that avoids collisions. This local path is inferred by following the arrival time gradients between the two nodes. In this way, global planning is seamlessly integrated with local neural motion planning, ensuring both efficiency and collision avoidance when navigating large indoor environments.

\begin{table}[h]
\centering
\scalebox{1}{ 
\begin{tabular}{ccccc}
\toprule
    Env           &               & Ours              & ANTFields         & TDM \\
    \midrule
                  & SR   & 100\%             & 97\%              & 98\%\\
    Rancocas (7)      & FT    & 1.88 $\pm$ 0.28   & 2.57 $\pm$ 0.12   & 2.75$\pm$0.59\\
                  & MT  & 118.76            & 144.25            & 154.64\\
    \midrule
                  & SR   & 98\%              & 96\%              & 96\%\\
    Cantwell (8)      & FT    & 2.08 $\pm$ 0.27   & 2.64 $\pm$ 0.10   & 2.23 $\pm$ 0.29\\
                  & MT  & 118.64            & 133.55            & 122.31\\
    \midrule
                  & SR   & 99\%              & 82\%              & 96\%\\
    Pleasant (11)      & FT    & 2.11 $\pm$ 0.17   & 3.17 $\pm$ 0.05   & 2.31 $\pm$ 0.16\\
                  &  MT  & 186.81            & 227.96            & 196.35\\
    \midrule
                  &  SR   & 93\%              & 66\%              & 75\%\\
    Maguayo (31)       &  FT    & 2.20 $\pm$ 0.18   & 3.72 $\pm$ 4.57   & 3.87 $\pm$ 0.19\\
                  &  MT  & 277.57            & 375.99            & 380.02\\
    
\bottomrule
\end{tabular}}
\caption{Quantitative comparison of our method against ANTFields \cite{antfield}, and TDM \cite{ni2025physicsinformed}. SR refers to the motion planning success rate after the environment is fully explored. FT is the training time per frame, and MT is the total mapping time. }
\label{table:mapping}
\vspace{-0.28in}
\end{table}


\begin{figure}[t]
\centering
\begin{subfigure}{0.49\textwidth}
\centering
\includegraphics[trim={0cm 0cm 0cm 0cm}, clip,width=0.8\linewidth]{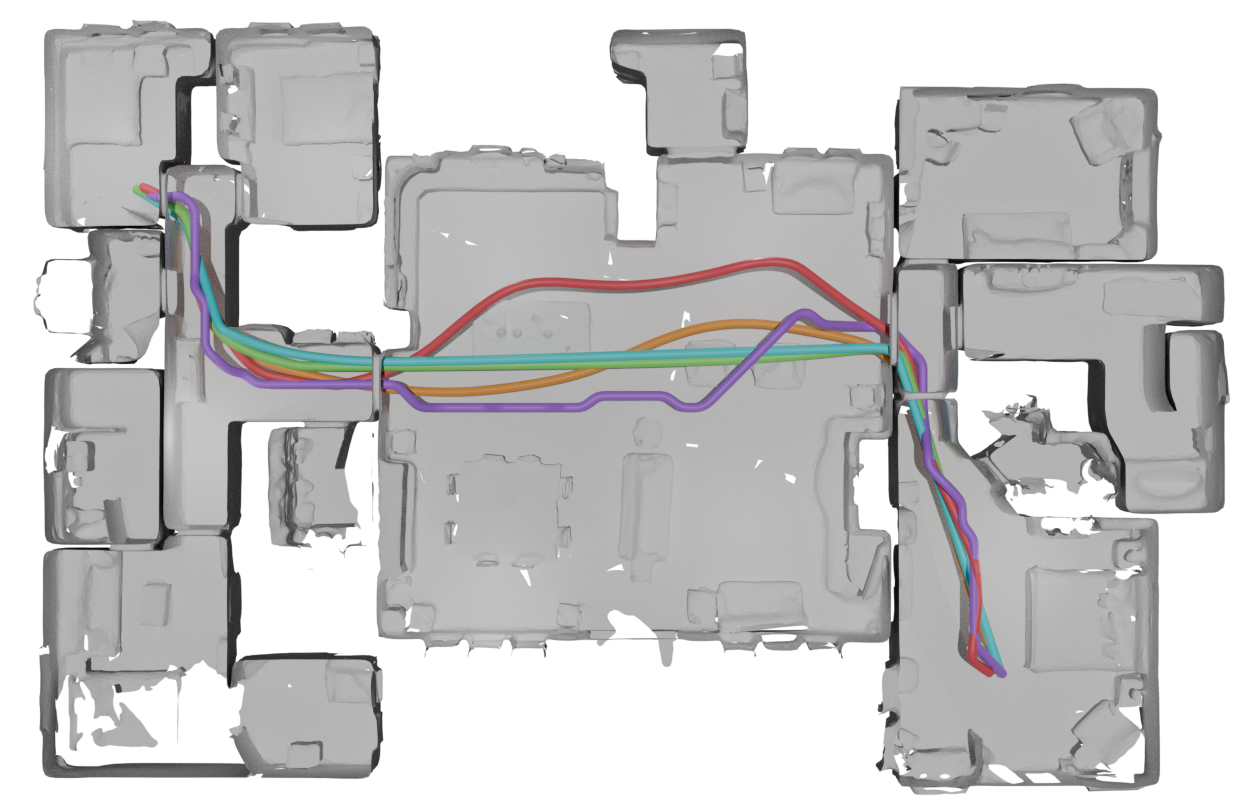}
\caption{{Sultan: 357.887 Sq. m}}
\end{subfigure}
\begin{subfigure}{0.5\textwidth}
\centering
\includegraphics[trim={0cm 0cm 0cm 0cm}, clip,width=0.64\linewidth]{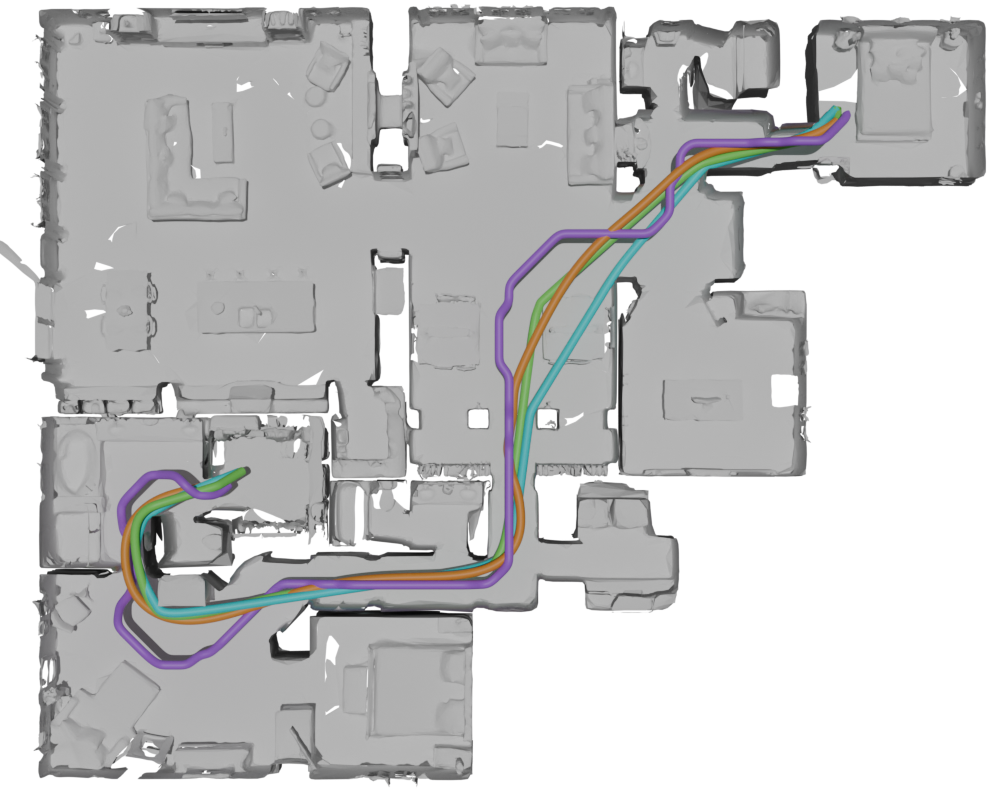}
\caption{Sanctuary: 289.285 Sq. m}
\end{subfigure}
\caption{Depiction of two Gibson environments: The paths generated by all methods are shown between the given start and goal. Our method successfully planned a collision-free smooth path in around 0.06 seconds, showcasing its ability to be deployed into complex indoor environments. SMP methods RRTConnect (Green), and Lazy-PRM (Cyan) took around 3 seconds to find the path, however with sharper turns. FMM (Purple) takes around 0.8 seconds to complete, but retrieved a longer path length due to limited discretization resolution. MPOT (Red) failed to retrieve a path in (b) due to correct path requiring multiple turns.}
\label{gibson}\vspace{-6mm}
\end{figure}
\begin{table*}[h!]
	\centering
	\begin{tabular}{ccclccccc}
		\specialrule{.15em}{.1em}{.1em}
		\textbf{Env. name} & \textbf{\# Rooms} & \textbf{Size (m$^2$)} &              & \textbf{Ours} & \textbf{MPOT}     & \textbf{FMM}    & \textbf{RRTConnect} & \textbf{LazyPRM}     \\ 
            \hline
		                   &                   &                       & Success rate & 100\%          & 91.0\%   & 100\%           & 100\%               & 100\%                      \\ 
		Rancocas          & 7                 & 116.738               & Time          & 0.08 $\pm$ 0.05 & 0.25 $\pm$ 0.04    & 0.79$\pm$ 0.01 & 0.44$\pm$ 0.61 & 1.06$\pm$ 1.07  \\
		                   &                   &                       & Path length  & 7.75 $\pm$ 4.21   & 9.00 $\pm$ 2.53  & 8.43$\pm$4.02   & 7.90$\pm$3.99       & 7.08$\pm$3.89     \\ 
            \hline
		                   &                   &                       & Success rate & 99.5\%        & 91.5\%       & 100\%           & 100.0\%             & 100\%                    \\ 
		Cantwell          & 8                 & 107.582               & Time & 0.07 $\pm$ 0.03         & 0.41 $\pm$ 0.21    & 0.82$\pm$0.02 & 0.61 $\pm$ 0.75       & 1.51$\pm$0.78     \\ 
		                   &                   &                       & Path length  & 8.80 $\pm$ 4.19  & 10.25 $\pm$ 2.80 & 9.03$\pm$4.48   & 8.83 $\pm$ 3.59     & 8.85$\pm$3.82     \\ 
            \hline
		                   &                   &                       & Success rate & 99.5\%          & 96.5\%  & 100\%           & 100.0\%             & 100\%                       \\ 
		Sultan             & 16                & 357.887               & Time         & 0.07 $\pm$ 0.05 & 0.57 $\pm$ 0.18  & 0.79$\pm$0.02   & 1.37 $\pm$ 2.08     & 1.51$\pm$0.75      \\ 
		                   &                   &                       & Path length  & 10.75 $\pm$ 6.67 & 15.27 $\pm$ 4.23  & 11.95$\pm$7.14  & 10.35 $\pm$ 5.98    & 10.67$\pm$6.12    \\ 
		\hline
		                   &                   &                       & Success rate & 99.0\%          & 96.5\%   & 100\%           & 100\%               & 100\%                      \\ 
		Sanctuary          & 19                & 289.285               & Time         & 0.07 $\pm$ 0.05  & 0.57 $\pm$ 0.18  & 0.81$\pm$0.02   & 1.37 $\pm$ 2.08     & 1.79$\pm$1.05     \\ 
		                   &                   &                       & Path length  & 10.38 $\pm$ 6.65 & 13.16 $\pm$ 3.67 & 11.40$\pm$8.08  & 10.68 $\pm$ 7.18    & 11.41$\pm$8.96    \\ 
		\hline
		                   &                   &                       & Success rate & 97.0\%         & 77.0\%   & 100\%           & 100.0\%             & 100\%                       \\ 
		Coronado           & 22                & 321.079               & Time         & 0.11$\pm$0.06  & 1.11 $\pm$ 0.31   & 0.77$\pm$0.02   & 1.55 $\pm$ 2.21     & 1.49$\pm$0.68      \\ 
		                   &                   &                       & Path length  & 14.99$\pm$9.30  & 13.81 $\pm$ 3.82  & 15.27$\pm$7.27  & 15.77 $\pm$ 10.34   & 14.21$\pm$5.38    \\ 
		\hline
		                   &                   &                       & Success rate & 99.0\%           & 65.0\%     & 100\%           & 100\%               & 100\%                   \\ 
		Keweenaw           & 24                & 314.952               & Time         & 0.13 $\pm$ 0.06  & 1.00 $\pm$ 0.28   & 0.79$\pm$0.01   & 1.75$\pm$1.93       & 1.07$\pm$2.04    \\ 
		                   &                   &                       & Path length  & 15.20 $\pm $0.42 & 15.46$\pm$8.27 & 16.71$\pm$8.70  & 18.07$\pm$9.85      & 15.58$\pm$8.29      \\ 
		\hline
		                   &                   &                       & Success rate & 93.5\%          & 11.5\%    & 100.0\%         & 100.0\%             & 100.0\%                   \\ 
		Frankton           & 30                & 1018.387              & Time         & 0.15 $\pm$ 0.05 & 0.74 $\pm$ 0.37  & 0.82$\pm$0.01   & 5.38 $\pm$ 3.81     & 1.79$\pm$1.00      \\ 
		                   &                   &                       & Path length  & 28.03 $\pm$ 15.45& 15.90 $\pm$ 3.19 & 27.59$\pm$12.79 & 29.07 $\pm$ 19.22   & 25.65$\pm$11.04   \\ 
		\hline
		                   &                   &                       & Success rate & 94.5\%           & 23.0\% & 100.0\%         & 100.0\%             & 100.0\%                     \\ 
		Leilani            & 42                & 550.098               & Time         & 0.12 $\pm$ 0.06  & 0.79 $\pm$ 0.39  & 0.80$\pm$0.01   & 5.33 $\pm$ 3.90     & 1.59$\pm$1.17     \\ 
		                   &                   &                       & Path length  & 20.93 $\pm$ 10.78 & 16.49 $\pm$ 5.38 & 19.02$\pm$9.83  & 19.39 $\pm$ 16.13   & 18.65$\pm$8.57    \\ 
		\specialrule{.15em}{.1em}{.1em}
	\end{tabular}
	\caption{Performance Comparison Across Environments. Our method achieved consistent success rate across environments while having the least amount of processing time. Sampling-based methods such as RRTConnect and LazyPRM achieve a higher success rate at the cost of increased processing time. FMM with a fixed discretization resolution retains a consistent success rate and processing time, however, with a path of longer length with many small turns. Optimization-based method, MPOT, fails on larger environments.}
    \label{baseline_comparison}
    \vspace{-5mm}
\end{table*}

\section{Experiments}

In this section, we present the experimental results of our proposed method. Since our approach builds upon ANTFields and TDM, we first compare its exploration and mapping performance to those methods in unknown environments. We then compare our approach's navigation capabilities against various methods to demonstrate its ability to generate fast, accurate paths in large, complex environments. Finally, we highlight the sim-to-real transfer of our method on a legged-robot by showing motion planning results in real-world settings with multiple rooms and long, narrow hallways.

\subsection{Mapping Effectiveness}

We begin by comparing our method against two alternative approaches for exploration tasks across four Gibson environments in different scales. The first approach, ANTFields, and the second, TDM, both lack the hierarchical and modular structure that characterizes our method. Note TDM is an offline method designed for known environments, and for a fair comparison, it utilizes the same training data from exploration as our approach. The aim of this comparison is to show that our exploration pipeline and hierarchical structure not only manage large, complex environments effectively but also achieve superior mapping accuracy and faster mapping times.

We evaluate our performance using three key metrics: planning success rate (SR), frame training time (FT) per observation, and total mapping time (MT). The planning success rate is calculated after exploration using 200 randomly sampled start-goal pairs in collision-free configurations. As shown in Table \ref{table:mapping}, our method achieves much higher success rates and faster exploration and training times in all environments. Although TDM improves the effectiveness of ANTFields, it still encounters challenges in larger environments and requires more training data to cover extensive areas with a longer per-frame training time. Both ANTFields and TDM exhibit longer mapping times due to slow convergence. In contrast, our integration of TDM within a hierarchical and modular framework results in more robust mapping and navigation, as demonstrated by our results.


\subsection{Motion Planning Effectiveness}

Next, we explicitly compare the navigation modules to illustrate our method’s advantages in planning speed and accuracy across environments ranging from 6 to 42 rooms. We include several state-of-the-art baseline methods. For sampling-based approaches, we evaluate RRTConnect\cite{rrtc}, a bi-directional sampling algorithm, and LazyPRM\cite{lazyprm}, which constructs a roadmap suitable for path planning. For grid-based methods, we use the Fast Marching Method (FMM)\cite{fomel2009fast}, which also solves the Eikonal equation via wave propagation as our method does. We also compare against MPOT\cite{mpot}, a state-of-the-art optimization-based planner. We use success rate, planning time, and path length as our metrics, and randomly select the same 200 start–goal pairs for each environment to ensure fairness. For RRTConnect and LazyPRM, we set a 10-second time limit, with any exceedance also counted as a failure. The average training time for our method in all environment is 18 seconds per network, where the number of network increasing approximately linearly with the number of rooms. The roadmap method LazyPRM has a preconstruction time in ranging from 22 seconds for the small environments, and 7 minutes for the largest environment. The planning time for LazyPRM listed in Table~\ref{baseline_comparison} does not include preconstruction time.  

Table~\ref{baseline_comparison} shows that the optimization-based method MPOT achieves fast inference times but is prone to local minima and struggles in complex environments. In contrast, classic sampling-based methods offer correctness guarantees at the cost of longer planning times, which often take seconds. FMM delivers both accuracy and speed; however, its discretized paths are still slower than ours and are sensitive to grid resolution. 

Regarding path length from Table~\ref{baseline_comparison} and path quality from Fig.~\ref{gibson}, MPOT typically produces suboptimal, longer paths in simple environments, whereas in more complex settings, it could only solve simple problems where the start and goal were not too far. Sampling-based methods yield variable path lengths due to random sampling. LazyPRM, for example, can find short paths with dense preconstructed samples, but may result in trajectories with sharp turns and near obstacles, as illustrated in Fig.~\ref{gibson}. FMM generally finds smooth paths by minimizing travel time, yet its grid-search nature often introduces multiple small turns, highlighting a trade-off between path quality and planning time.

In contrast, our method solves the Eikonal Equation to compute the shortest travel time and, due to its implicit representation, is capable of generating smooth trajectories for any start-and-goal pair. Overall, our approach produces high-quality, correct paths with fast planning times and scales effectively to large, complex environments.






\subsection{Real demonstration}

Finally, we demonstrate the real-world effectiveness of our method by deploying it in large, complex environments. Specifically, we tested in two distinct real-world setups, designated as Environment A (1883.39 m$^2$) and Environment B (845.65 m$^2$). Environment A shown in Fig~\ref{fig:lwsn_basement} comprises four classrooms connected by a common walkway, while Environment B features a long hallway. For both setups, we used a Unitree B1 quadruped robot equipped with a PandarXT-16 LiDAR sensor for perception. Figure \ref{fig:lwsn_basement} shows the robot dog navigating Environment A. Our room segmentation algorithm successfully identified the individual rooms and split the hallway into two smaller, more manageable sections. The path planning process was completed in 0.08 seconds, while training took around 3 minutes. Video demonstrations of the robot navigating these environments using our method can be found in the supplementary material.


\section{Conclusions and Future Work} 
\label{sec:conclusion}

In this paper, we introduced Modular-NTFields (mNTFields) that utilize neural Eikonal PDE solvers to learn a hierarchical navigation planning policy in unknown, large environments, without the need for expert demonstration trajectories. Our experiments demonstrate that mNTFields outperforms previous Eikonal PDE solvers when mapping expansive, unfamiliar environments. Additionally, it achieves faster and more accurate planning than prior navigation methods in indoor settings with 6 to 42 rooms. We further validate the real-world performance of our approach by showing that it scales effectively in real-world scenarios, including long, narrow hallways and multiple doors.

For future work, we plan to explore hardware-accelerated tools to support the parallel training of local planners. We also aim to investigate the applicability of our method in dynamic environments and assess its performance in higher-dimensional planning tasks.






\bibliographystyle{IEEEtran}
\bibliography{references}

\end{document}